%% file: m2909.tex
\documentclass{ecai} 

\usepackage{latexsym}
\usepackage{amssymb}
\usepackage{amsmath}
\usepackage{amsthm}
\usepackage{booktabs}
\usepackage{enumitem}
\usepackage{graphicx}
\usepackage{color}

\usepackage[ruled,vlined,linesnumbered,noresetcount]{algorithm2e}   
\usepackage{multirow}

\newcommand{\genemask}{\textsc{GeneMask}}
\newcommand{\globalgenemask}{\textsc{Global}}
\newcommand{\dyngenemask}{\textsc{CM-GEMS}}

\newcommand{\BibTeX}{B\kern-.05em{\sc i\kern-.025em b}\kern-.08em\TeX}

\begin{document}


\begin{frontmatter}


\paperid{2909} 


\title{Unlocking Efficiency: Adaptive Masking for Gene Transformer Models}


\author[A]{\fnms{Soumyadeep}~\snm{Roy}\thanks{Corresponding Author. Email: soumyadeep.roy9@iitkgp.ac.in Accepted for publication at the 27th European Conference on Artificial Intelligence ECAI 2024. This is the author's version of the work. It is posted here for your personal use. Not for redistribution.}}
\author[A]{\fnms{Shamik}~\snm{Sural}}
\author[A]{\fnms{Niloy}~\snm{Ganguly}}

\address[A]{Indian Institute of Technology Kharagpur}

\input{abstract}

\end{frontmatter}

\input{introduction}
\input{related_works}
\input{building-blocks}
\input{method}
\input{experiments}
\input{conclusion}

\begin{ack}
Soumyadeep Roy is supported by the Institute Ph.D. Fellowship at the Indian Institute of Technology Kharagpur.
\end{ack}



\bibliography{m2909}

\newpage

\appendix
\input{appendix}

\end{document}

%% file: abstract.tex
\begin{abstract}
Gene transformer models such as Nucleotide Transformer, DNABert, and LOGO are trained to learn optimal gene sequence representations by using the \textit{Masked Language Modeling} (MLM) training objective over the complete Human Reference Genome. However, the typical tokenization methods employ a basic sliding window of tokens, such as k-mers, that fail to utilize gene-centric semantics. This could result in the (trivial) masking of easily predictable sequences, leading to inefficient MLM training. Time-variant training strategies are known to improve pretraining efficiency in both language and vision tasks. In this work, we focus on using curriculum masking where we systematically increase the difficulty of \textit{masked token prediction} task by using a \textit{Pointwise Mutual Information}-based difficulty criterion, as gene sequences lack well-defined semantic units similar to words or sentences of NLP domain. Our proposed \textit{Curriculum Masking-based Gene Masking Strategy} (\dyngenemask) demonstrates superior representation learning capabilities compared to baseline masking approaches when evaluated on downstream gene sequence classification tasks. We perform extensive evaluation in both few-shot (five datasets) and full dataset settings (\textit{Genomic Understanding Evaluation} benchmark consisting of 27 tasks). Our findings reveal that \dyngenemask{} outperforms state-of-the-art models (\textit{DNABert-2, Nucleotide transformer, DNABert}) trained at 120K steps, achieving similar results in just 10K and 1K steps. We also demonstrate that Curriculum-Learned LOGO (a 2-layer DNABert-like model) can achieve nearly $90\%$ of the state-of-the-art model performance of 120K steps. We will make the models and codes publicly available at \url{https://github.com/roysoumya/curriculum-GeneMask}.
\end{abstract}

%% file: introduction.tex
\section{Introduction}
The field of developing large foundational models in genomics is gaining popularity ~\citep{consens2023transformers}; these models are primarily trained on the Human Reference Genome, such as DNABert, LOGO, and Nucleotide Transformer (Human Reference Genome model variant). These models use Masked Language Modeling (MLM), pioneered by BERT~\citep{bert}, where the model predicts original tokens of a masked subset as clues. The state-of-the-art pretrained models (DNABert~\cite{DNABert} and LOGO~\cite{yang:2021:bioarxiv:logo}) in gene sequence classification tasks such as \textit{promoter region prediction, core promoters, enhancers, splice sites prediction, and functional genetic variants}, are widely used in literature~\cite{Badirli2021,genebert2021}. The recently introduced \textit{Genome Understanding Evaluation}~\citep{zhou2024dnabert} (GUE) benchmark covers these gene sequence modeling tasks for multiple species such as \textit{human, mouse, yeast,} and \textit{virus}; this serves as a standard resource for evaluating the gene foundational models. Here, each task takes as input a sequence of nucleotides that consists of bases such as adenine (A), cytosine (C), guanine (G), and thymine (T), and an associated class label. The gene transformer models are first pretrained using the masked language modeling (MLM) objective to learn optimal gene sequence representations. However, unlike the NLP domain, where well-defined semantic units called words or sentences exist, no well-defined semantically demarcated tokens exist for a given sequence.

\begin{figure}[b]
    \centering
    \includegraphics[width=0.44\textwidth]{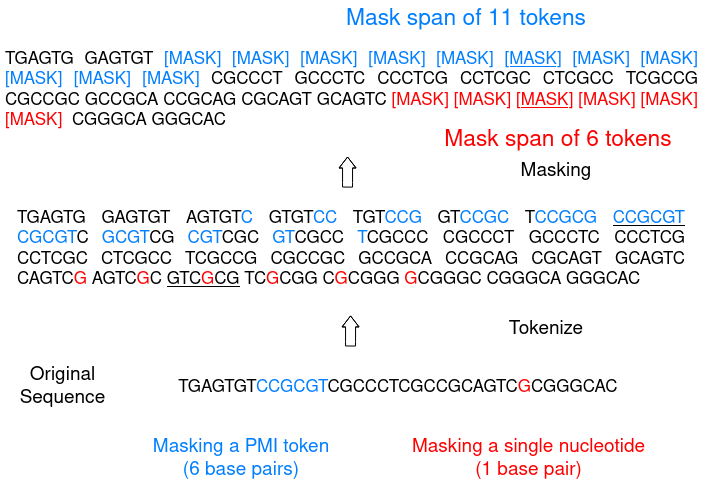}
    \label{fig:kmertokenization}
            \caption{Masked language modeling pipeline of gene transformer models. The original sequence is broken into k-mers (here, k $= 6$). The difference between masking a single nucleotide versus a PMI token is highlighted by using red and blue font, respectively}
\end{figure}

Therefore, to come to a workable solution, researchers randomly select a sequence of $k$ nucleotides~\cite{DNABert,genebert2021}; we use $k = 6$ throughout this study, following prior works~\citep{royecai2023,DNABert,yang:2021:bioarxiv:logo}. Figure~\ref{fig:kmertokenization} clearly depicts converting a gene sequence into a sequence of k-mers. For example, if `CCGCGT' is a frequently occurring sequence, then if we mask just one nucleotide within, it may be easily predicted based on its highly co-occurring nucleotides. Therefore, the model can easily predict the masked token by learning trivial patterns, leading to increased pretraining time. DNABert requires a total time of $25$ days to complete pretraining for 120K steps on $8$ NVIDIA 2080Ti GPUs.~\citet{consens2023transformers} mentioned this requirement of huge compute as a key limitation of current gene foundational models, and approximates the cost of running the popular models of DNABERT~\citep{DNABert}, Enformer~\citep{enformer}, Nucleotide Transformer~\citep{dalla2023nucleotide} and HyenaDNA~\citep{hyenadna}, at a cost of \$5000 USD on 8 A100 GPUs. Therefore, we believe efficient and fast pretraining of gene foundational models is an important research area. This work focuses on fast pretraining of gene transformer models such as DNABert and LOGO.

~\citet{anagnostidis2023navigating} discusses the concept of `compute-optimal' training and shows that adaptive training strategies significantly reduce the required compute to reach a given target performance compared to their static counterparts. We aim to achieve equivalent performance for a specified model (gene transformers in our case) with fewer computational resources than initially projected. We can develop an efficient pretraining strategy whereby we mitigate the instances of `easy' learning by masking entire spans of highly correlated spans that occur frequently. This prevents the model from wasting its pretraining steps on predicting such trivial cases. 
 Such intelligent masking schemes~\citep{pmi-masking,spanbert2020} have been popular in the NLP domain. ~\citet{royecai2023} developed the \textsc{GeneMask} masking strategy and showed such a reduction in the number of pretraining steps, from 120K steps to 10K steps in the case of DNABert. This paper shows two reduction schemes: (i) further reduction in pretraining steps - 10K to 1K, and (ii) reduction in model size in terms of layers or transformer blocks. (12 to 2). 

 \citet{royecai2023} utilize the Pointwise Mutual Information (PMI) score to prioritize highly correlated spans within a gene sequence instead of the random span masking strategy as used in the case of DNABert (or LOGO). PMI is computed as the probability of a set of tokens occurring together compared to appearing independently throughout the corpus. A high PMI score indicates a high correlation. Specifically, \textsc{GeneMask}~\citep{royecai2023} first randomly selects certain positions as mask centers and then selects the highest-ranked PMI token present within a fixed neighborhood of that mask center. 
However, this strategy potentially misses the top-ranked PMI tokens in a gene sequence and may instead mask lower-ranked PMI tokens, thus not fully utilizing the PMI information. This may allow further wastage of pretraining steps and thus result in a delayed convergence. To mitigate this issue, we develop \globalgenemask{} masking strategy that always captures the top-m highest ranked PMI tokens in a sequence, and thus ensures that the PMI information is strictly used, and may potentially lead to faster convergence and better few-shot performance. We observe that \globalgenemask{} outperforms all the baseline models at 10000 steps when tested in the few-shot setting (combined 10 and 50-shot).

Few studies advocate using dynamic pretraining strategies, which is slowly garnering research interest. \citet{yang-etal-2023-learning} is the first work to explore time-variant MLM settings into account; they develop two strategies called \textit{Masking Ratio Decay} and \textit{POS-Tagging Weighted Masking}, where they adaptively tune the masking ratio and masked content as the pretraining progresses. Combining the idea of curriculum masking, we use PMI to regulate/decide the difficulty level, whereby during the pretraining step, tokens for easier concepts are masked first. In contrast, in the later part, the harder concepts are used. Therefore, contrary to the time-invariant masking strategy of \textsc{GeneMask} (i.e., stays fixed during the entire pretraining procedure), we develop a dynamic (time-variant) masking strategy that updates the masking algorithm (and consequently the masking ratio) during pretraining. Our proposed method \dyngenemask{} achieves better few-shot performance than competing baseline models over the five standard gene sequence classification datasets (gene regulatory elements such as promoters, enhancers, silencers, and splice sites). We also perform extensive ablation analysis and experiments to better understand the efficacy of \dyngenemask{}. We make the codes and data publicly available at \url{https://github.com/roysoumya/curriculum-GeneMask}. 

In this work, we make the following contributions:
\begin{enumerate}
    \item To the best of our knowledge, \dyngenemask{} is the first work to apply curriculum masking and time-variant (dynamic) pretraining strategy for training gene transformer models.  \dyngenemask{} trained for 10K steps outperforms the state-of-the-art models trained for 120K as well as the state-of-the-art gene masking algorithm, \genemask{}, by $2.18\%$ and $1.24\%$ respectively on average across both `Human' and `non-Human' species of the GUE benchmark~\citep{zhou2024dnabert}. 
    \item We develop the \globalgenemask{} masking strategy that directly (globally) selects the top-ranked PMI tokens throughout the entire gene sequence (of 512 tokens) and outperforms \genemask{} by $2.66\%$ on average across five datasets and their corresponding 10 and 50-shot settings.
    \item We perform extensive evaluation over 32 datasets (5 few-shot and 27 full training data) across two gene transformer model architectures of DNABert and LOGO. Although Gene Mask showed improvement only in terms of few-shot performance, we also show results on the full dataset (GUE benchmark) and observe that \dyngenemask{} at 1K steps and \globalgenemask{} at 10K can achieve performance within a $90\%$ margin than state-of-the-art models such as DNABert-2 and Nucleotide transformers trained for 120K steps.
\end{enumerate}

%% file: related_works.tex
\section{Related Works}
Here, we summarize the recent works on gene sequence modeling, followed by various masking strategies used for masked language modeling (MLM) based pretraining strategies, including time-variant pretraining and curriculum masking. 

\noindent \textbf{Representation learning for genomics.}
Badirli et al.~\cite{Badirli2021} represent the DNA barcodes as a vector representation for fine-grained species classification. Gene sequences are typically represented as k-mers by recent works~\citep{dnavec,DNABert,yang:2021:bioarxiv:logo}, which is then used to learn a dense representation from an adapted BERT model. However, other works such as BigBird~\citep{bigbird2020} and DNABert-2~\citep{zhou2024dnabert} use a \textit{byte-pair encoding} scheme to build the model vocabulary.~\citet{genebert2021} develop a multimodal pretrained model using a combination of gene sequences and transcription factors. There is a recent line of work to learn long-range interactions present in the human genome: BigBird~\citep{bigbird2020}, Enformer~\citep{enformer} and HyenaDNA~\citep{hyenadna}. Given the context length limitation of transformer architectures, ~\citet{enformer} utilizes a combination of convolutional and transformer layers to encode the long sequences of one-hot encoded base pairs into a more condensed vector representation. However, the \textit{Enformer} model focuses on gene expression track prediction, which is not the focus of this work. In this work, we focus on gene sequence classification tasks. 

\noindent \textbf{MLM masking strategies in NLP.} 
BERT~\citep{bert} apply \emph{random token masking} where $15\%$ of the input tokens are chosen to be masked uniformly. Whole word masking~\citep{senrich:2016:acl:wholewordmasking} and Entity masking~\citep{sun2019ernie} show performance improvement over the standard `random token masking' scheme. Here, they masked a contiguous sequence of tokens that either formed a whole word (\textit{whole word masking}) or an entity (\textit{entity masking}). However, a more straightforward approach called `random span masking' outperformed entity masking, where the mask centers are randomly selected, and the span length w.r.t this mask center is drawn from a geometric distribution. However, there are no pre-defined semantics similar to entities and words of NLP, which can be applied to gene sequences; therefore, it is impossible to apply entity and whole-word masking in our case. ~\citet{pmi-masking} use the Pointwise Mutual Information (PMI) score to identify the sequence of contiguous tokens to be masked together. Such intelligent masking shows accelerated performance whereby they achieve end-of-pretraining performance at almost half the training steps. ~\citet{sadeq-etal-2022-informask} proposes the `Informative Relevance' metric, which, for a given word, is computed as the sum of PMI values between a masked word and all unmasked words in the given sentence. However, the same issue is the lack of gene semantics, where the equivalent of a natural language sentence is not known for the gene sequence modeling domain. 

 \noindent \textbf{Time-variant pretraining and curriculum masking.} Curriculum learning is also used in reinforcement learning framework~\citep{yasutomi2023}. Curriculum masking easy-to-hard schedule for masked image modeling using vision transformer blocks~\citep{Madan2024WACV}. ~\citet{pavlova-makhlouf-2023-bioptimus} used easy-to-hard curriculum masking for biomedical named entity recognition tasks. The notion of task difficulty involves (i) masking words instead of tokens and (ii) increasing the masking rate. \citet{yang-etal-2023-learning} demonstrate that using time-invariant pretraining, particularly in terms of masking ratio and masked content, does not produce good results. They propose a novel masked content selection strategy based on the parts of speech of a given word; masking non-function words proved to be more useful than masking function words.~\citet{lee-etal-2022-efficient-pre} introduced the idea of curriculum masking (easy-to-hard) where they first map words or phrases to a concept in an existing knowledge graph called Concept-Net. They define words and phrases related to many other concepts as `easy concepts' and use them to initialize the curriculum masking schedule. As the pretraining progresses, it gradually masks concepts related to the previously masked concepts during the consecutive stages.  
 
 However, in our case, we neither have any word or phrase-level gene semantics nor access to such a large, well-annotated knowledge graph.~\citet{lee-etal-2022-efficient-pre} applied it in the NLP domain where they evaluated the General Language Understanding Evaluation (GLUE) benchmark that comprises single sentence or sentence pair classification tasks. ~\citet{lee-etal-2022-efficient-pre} found that the hard-to-easy curriculum performs poorly and leads to a drop in performance, which they explain may be because the hard concepts at the start proved to be too difficult without prior knowledge or learning from relevant, easier concepts. Therefore, in our proposed time-variant masking strategy, \dyngenemask{}, we initially adopt an easier masking strategy of \genemask{}, followed by a hard masking strategy, \globalgenemask{} that only masks the top-ranked PMI tokens.

%% file: building-blocks.tex
\section{Research Background}\label{sec:dnabert-config}
Here, we introduce the two popular gene transformer models of DNABert~\cite{DNABert} and LOGO~\cite{yang:2021:bioarxiv:logo} that are based on the BERT architecture. These models learn a robust contextual representation of DNA sequence fragments by leveraging unsupervised training based on Masked Language Modeling over the Human Reference Genome; it contains 3.2 billion nucleotides over 24 chromosomes. We will explain the complete pipeline by mapping to the stages similar to pre-trained language models (PLMs) developed for natural language processing tasks. We divide it into two phases: (i) preprocessing and (ii) pretraining. We use the same experimental design and training strategy as used by~\citet{royecai2023} for evaluating the different masking strategies for fast pretraining of gene transformer models.

\subsection{Preprocessing}
\noindent \textbf{Tokenization of gene sequences.} Here, we represent a gene sequence as a sequence of k-mers, as used by prior studies~\cite{DNABert,dnavec} while developing gene transformer models. $k$-mer represents a sliding window of length $k$, as depicted in Figure~\ref{fig:dyngenemask}. We adopt the value of $k$ as $6$, following the setup used by~\citet{royecai2023}. 6-mers are more appropriate because they incorporate richer contextual information and have a manageable memory and compute requirement~\citep{yang:2021:bioarxiv:logo}.

\noindent \textbf{Pretraining data preparation.} We use the pretraining data constructed and made available by~\citet{royecai2023} to maintain fair comparison. Next, we summarize their pretraining data construction steps. We obtain the \textit{Human Reference Genome} data in the \textit{FASTA} format from the \textit{Genome Reference Consortium Human Build 38 patch release 13} (GRCh38.p13)~\cite{human-reference}. This dataset comprises long, unlabeled gene sequences divided across multiple chromosomes and is used by gene transformer models to perform unsupervised training using masked language modeling loss. We do not allow sequences that contain bases apart from that of A, T, C, or G. The sequence length is determined as $510$ token long for half the cases and a random length between 5 and 510 for the remaining half.

\subsection{Pretraining}
The standard masked language modeling (MLM) loss, the same as that used for BERT~\citep{bert}, is used as a pretraining objective for training DNABert and LOGO. A key difference between the NLP domain and gene transformers is that to mask a single nucleotide, one has to mask a continuous sequence of tokens to avoid trivial masked token prediction. This is because a single nucleotide is present as part of $k$ contiguous k-mers. 
In a more formal context, let us denote a nucleotide as DNA[$i$] while representing a 6-mer token as T[$i$], which is equivalent to {DNA[$i$-2], DNA[$i$ - 1] $\cdots$ DNA[$i$ + 3]}. Consequently, the tokens T[$j$], $\forall(j)_{j=i - 2}^{i + 3}$ are subject to masking. Given a token length of $k=6$ and the requirement to mask $15\%$ of tokens~\cite{bert}, the probability for masked language model (MLM) is established at $15\% / 6$, resulting in $2.5\%$ of nucleotides chosen for masking (given our primary focus on 6-mers unless specifically indicated as T[$i$] = $T_6[i]$).

\subsection{\genemask{}: State-of-the-art Masking Strategy}\label{sec:genemask-strategy}
 \genemask{}~\citep{royecai2023} is the state-of-the-art time-invariant masking algorithm for Masked Language Model (MLM) training of gene transformer models. It showed improvement over the random span masking strategy used by DNABert and LOGO by masking all the nucleotides simultaneously in the most correlated spans (determined using \textit{Pointwise Mutual Information value, PMI}~\citep{pmi-masking}). Masking correlated spans helps to reduce information leakage and enhances the model's ability to predict intricate patterns. However, retaining the advantages of the conventional random masking approach is also necessary. Therefore, \genemask{} first randomly selects positions within a gene sequence as mask centers, which is then followed by locally selecting top-ranked \textit{normalized PMI} (NPMI\textsubscript{k}, see Equation~\ref{eq:norm-pmi}) within a fixed neighborhood of the mask center, for masking purposes. The formal definition of Normalized Point-wise Mutual Information (NPMI\textsubscript{k}), where k $>$ 2, as proposed by~\citet{royecai2023}. 
{\small{
\begin{equation}\label{eq:norm-pmi}
	\textrm{NPMI}_k(w_1\ldots w_k) = \newline PMI_{k} * \frac{\log f(w_1\ldots w_k)}{\log (c)+ \log f(w_1\ldots w_k)} 
	\end{equation}
}}
\raggedbottom
 $f(w_1\ldots w_k)$ represents the number of times the k-mer sequence of $w_1\ldots w_k$ occurs. The minimum frequency of occurrence is referred to as $c$, which is used to filter rare tokens. 

%% file: method.tex
\section{Proposed Methodology}
Our proposed masking strategy for fast pretraining is called \textit{Curriculum Masking-based Gene Masking Strategy} (\dyngenemask). Optimizing the Pointwise Mutual Information (PMI) score prioritizes the masking of $k$ (=6)-mers that co-occur more frequently than their components. It takes full advantage of the existing \genemask{} strategy (Section~\ref{sec:genemask-strategy}). It later shifts to our proposed \globalgenemask{} masking strategy (Section~\ref{sec:global-pmi-strategy}) towards the end of the pretraining stage. Our key innovation lies in adopting a time-variant masking strategy that automatically transitions to a new masking strategy during the pretraining process when certain conditions (in this case, perplexity score) are satisfied. At the start of the pretraining, all tokens are unfamiliar to the models. However, as the pretraining progresses, the relative predictive difficulty of each token will vary. We also adopt an easy-to-hard curriculum masking schedule based on consistent observations of prior works~\citep{lee-etal-2022-efficient-pre,yang-etal-2023-learning}.

\subsection{Global PMI Masking Strategy}\label{sec:global-pmi-strategy}
We always mask the highest PMI tokens instead of locally optimal PMI tokens as done for \genemask. The \genemask{} masking strategy of random masking for $50\%$ of cases and locally optimal PMI masking for the remaining $50\%$ of cases is replaced by masking only the highest-ranked PMI tokens. Specifically, we select the top $m$ nucleotides as mask centers (MC) with the highest $NPMI_k$ score.  We term this masking strategy as `Global Masking Strategy' (\globalgenemask), and it is further explained in detail in Algorithm~\ref{algo:global-pmi-mask}.

\noindent \textbf{Determining the value of \textit{m}:} A PMI token corresponds to one 6-mers, which contains six nucleotides. Therefore, we need to mask a span of $11$ adjacent tokens to mask one PMI token because of six mask centers, two left-side tokens, and three tokens towards the right. The original MLM probability for masking one nucleotide was $15\% / 6 = 2.5\%$. In our case, it is gets reduced to $15\% / 11 = 1.36\%$. Given the input sequence length is $512$ tokens, $m$ $\approx$ 7.

\newcommand\mycommfont[1]{\footnotesize\ttfamily\textcolor{blue}{#1}}
\SetCommentSty{mycommfont}
\SetKwInput{KwInput}{Input}      
\SetKwInput{KwOutput}{Output}    
\SetKwInput{KwData}{Initialization}

\LinesNumberedHidden

\begin{algorithm}[!ht]
\footnotesize
\caption{\globalgenemask{} Algorithm}\label{algo:global-pmi-mask}

\DontPrintSemicolon \;
    \KwInput{Input sequence of 6-mer tokens having a maximum of 510 tokens, Pre-computed Normalized PMI\textsubscript{k} (NPMI\textsubscript{k}) values for all 6-mers stored as a dictionary} 
    \KwOutput{$MaskTokenSet$: Token indices within input sequence to be masked}
    \KwData{
    \tcp{A 6-mer token present at $i$-th position is represented as T[i], the $i$-th nucleotide is represented as DNA[i] }

    $MaskTokenSet \gets \emptyset$ 
    
    T[$i$] $\gets$ \{DNA[$i$-2]~$\cdots$~DNA[$i$+3]\}
    }
	\SetKwFunction{FMain}{MapNucleotideToKmerTokens}
	\SetKwProg{Fn}{Function}{:}{}
	
	\Fn{\FMain{nucleotide position id $i$}}{
          
            $MappedTokens$ $\gets$ T[$j$],  $\forall(j)_{j=i - 2}^{i + 3}$ \;
			
			\KwRet\  $MappedTokens$ \; 
    }
    
    \textbf{Step 1:} Sort the DNA string with 6-mer tokens in a non-increasing order of $NPMI_k$ score. 

    \textbf{Step 2:} Create a priority set with the top-ranked $m$ nucleotides with the highest $NPMI_k$ score.

    \textbf{Step 3:} \For{each nucleotide in priority set}	{
        \tcp{Masking a PMI token involves masking 11 adjacent tokens}
        $MaskTokenSet$ $\gets$ $MaskTokenSet$ $~\cup$ $\forall(j)_{j=\tau - 2}^{\tau + 3}$ $MapNucleotideToKmerTokens$( j )
     }
            
     \KwRet\  $MaskTokenSet$ \;
\end{algorithm}

\subsection{\dyngenemask: Curriculum Masking-based Gene Masking Strategy}
 \noindent \textbf{Motivation.} The \genemask{} masking strategy limits its search for the highest PMI tokens within the local context of the randomly selected mask centers. Although this enforces the PMI tokens being uniformly selected, \genemask{} may miss the top-ranked PMI tokens globally, i.e., within the entire sequence of 512 tokens. The top-ranked PMI tokens are the most difficult to predict and force the model to learn deeper patterns that eventually lead to faster convergence or pretraining. Since \genemask{} may miss masking the top-ranked PMI tokens, it may lead to trivial masking and delayed convergence. We develop \dyngenemask{} to address this limitation in this work. 

\begin{figure}[b]
    \centering
    \includegraphics[width=0.48\textwidth]{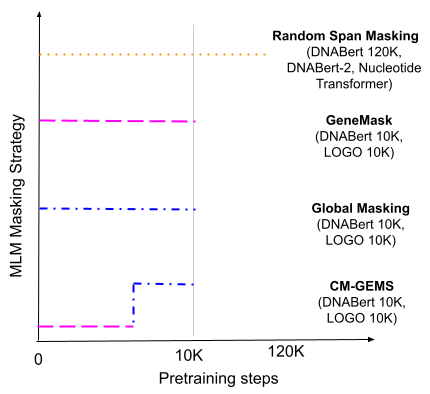}
    \label{fig:dyngenemask}
            \caption{Comparison of \dyngenemask{} and time-invariant strategy of~\globalgenemask{} with existing masking strategies}
\end{figure}

\noindent \textbf{Proposed Masking Algorithm of \dyngenemask.} \dyngenemask{} takes advantage of both \genemask{} algorithm proposed by~\citet{royecai2023} as well as the \globalgenemask{} algorithm proposed in this paper. We formulate it as a two-stage easy-to-hard curriculum masking strategy, where \genemask{} acts as the `easy' stage of the curriculum. We allow the \genemask{} masking strategy till the model continues to learn, which we measure by the drop in perplexity score. Once the drop in perplexity score falls below a threshold value (in our case, the value is one), we consider that the model improves marginally from this stage onwards. Therefore, we then shift to the \globalgenemask{} masking strategy, which is the `hard' stage of the curriculum, as it only considers the top-ranked PMI tokens. The only associated change to the pretraining hyper-parameters, apart from the masking algorithm, is the change in masking rate, which is controlled by the `mlm\_probability' field; it reduces from $1.765\%$ of \genemask{} to $1.36\%$ of \globalgenemask{}. Figure~\ref{fig:dyngenemask} depicts the two-stage curriculum masking strategy of \dyngenemask{}. To the best of our knowledge, this is the first work to explore the time-variant pretraining strategy for fast pretraining of genome foundational models.

%% file: experiments.tex
\section{Experimental Setup}
We describe the datasets and the evaluation setup, followed by the training details for~\dyngenemask{} and baseline models. 

\subsection{Datasets} 
\noindent \textbf{Full dataset evaluation.} We use the Genome Understanding Evaluation (GUE) benchmark recently introduced by DNABERT-2~\cite{zhou2024dnabert} for evaluating in the full dataset setting. It consists of $7$ gene sequence classification tasks with $28$ datasets with input sequence lengths ranging between 70 to 1000; we could not evaluate on \textit{Covid Variant Classification} task (one dataset) as it was not included in their evaluation scripts. Therefore, we finally tested on $27$ datasets, where the number of target classes is two (binary classification) except for \textit{Splice Site Detection}, which has three target classes. Unlike the few-shot evaluation setup described above that only focuses on human species, $15$ out of the $27$ datasets of GUE belong to Non-human species ($5$ from Mouse for the \textit{Transcription Factor Prediction} task and $10$ from Yeast for the \textit{Epigenetic Marks Prediction} task). This is an important additional feature that further helps us evaluate the multi-species transferability in the `genome understanding' capability of each model.

\noindent \textbf{Few-shot evaluation.} We evaluate across five gene sequence classification tasks, where the input is a gene sequence, i.e., a sequence of nucleotides, and the task is to predict whether it is a gene regulatory element or not. The promoter region prediction tasks (Prom-core, Prom-300) and Enhancer-Cohn prediction task (Cohn-enh) are obtained from~\citet{royecai2023}. We constructed the `silencer' dataset: the positive data points are obtained from the `SilencerDB' database~\cite{silencer2020}, and used the `high-throughput Homo-sapiens' version. We followed the same negative set construction technique as done for Prom-core and Prom-300, as well as for DeePromoter~\citep{deepromoter2019}.

\begin{table}[t]
    \centering
    \begin{tabular}{ccp{1cm}p{1cm}c}
    \hline
         \textbf{Dataset}&  \textbf{Task}& \textbf{Num. of classes}& \textbf{Sequence length}& \textbf{Train / Test}\\ \hline
         Prom-core&  Promoter& 2 &  70 & 53276 / 5920\\
         Prom-300&  Promoter& 2 &300& 53276 / 5920\\
         Cohn-enh&  Enhancer& 2 &500& 20843 / 6948\\
         Splice-40 & Splice Sites & 3&40 & 24300 / 3000 \\
         Sil-300&  Silencer&  2&300& 14909 / 1657\\ \hline
    \end{tabular}
    \label{tab:dataset-stats}
    \caption{Descriptive statistics of few-shot evaluation datasets}
\end{table}
\vspace{2em}

\subsection{Evaluation Setup}
We use the same metric for performance comparison, as mentioned in the respective prior works, to maintain a fair comparison. We report the statistical significance results based on paired t-test for the performance improvement by \dyngenemask{} over baseline models. 

\noindent \textbf{Full dataset evaluation on GUE.} We use the \textit{Matthews Correlation Coefficient} (MCC) evaluation metric as used by~\citet{zhou2024dnabert}. \textit{MCC} is used to measure the binary classification quality where it takes into account true and false positives and negatives. `MCC' can take values between $-1$ and $1$, where $+1$ indicates an exact match, and $-1$ represents a complete disagreement between the ground truth and prediction. 

\noindent \textbf{Few-shot evaluation.} We report accuracy as used by~\citet{royecai2023} for few-shot performance comparison of gene sequence classification tasks. The experiments are repeated for ten times by selecting a random seed and a set of $n$ data points for each class, corresponding to each run; the mean and standard deviation values are reported. We only show results for values of $n = 10, 50$, i.e., the 10-shot and 50-shot settings respectively. We do not perform any hyperparameter tuning as access to a validation dataset is not assumed, following the prior few-shot classification works~\citep{schick-schutze-2021-just,schick-schutze-2021-exploiting,royecai2023}.

\subsection{Training Details}
\noindent \textbf{Full dataset evaluation on GUE.} We directly use the evaluation script provided by~\citet{zhou2024dnabert} in their GitHub codebase 
for evaluating our proposed models. We use the performance scores reported by the authors of DNABert-2 and the Nucleotide transformer.

\noindent \textbf{Few-shot evaluation.} We follow the same training details as the work of ~\citet{royecai2023} for setting the GeneMask, ORI10K results, i.e., we use the same hyperparameter values as the original state-of-the-art (SoTA) models, except for `warmup steps' and `maximum pretraining steps' (see Section 5 of~\citet{royecai2023} to know more about training and fine-tuning) details, as well as the hyperparameters used).
We follow the same setup for the remaining model variants for the different pretraining step variants at $1000$ and $10000$.

\subsection{Baseline Models} 
We evaluate the impact of the different masking strategies on two gene transformer architectures of DNABert~\citep{DNABert} and LOGO~\citep{yang:2021:bioarxiv:logo}.

\noindent \textbf{State-of-the-art (SoTA) Models.} The recent gene transformer models such as DNABERT-2~\citep{zhou2024dnabert} and Nucleotide transformers~\citep{dalla2023nucleotide} do not make their pretraining codes publicly available, making it not possible to test the impact of our proposed masking strategy on such models. For a fair comparison, we select variants of SoTA models that are trained only on the Human Reference Genome and use the k-mer tokenization scheme (k=6). `SoTA-Best' represents the best result among the three k-mer-based SoTA models, such as DNABert 120K, NT-500M-Human, and DNABert-2 120K trained on 120K steps.

\noindent \textbf{Time-invariant Masking Strategies.} The models are pretrained for 10K steps using the same hyperparameter settings as~\citet{royecai2023} to obtain a fair comparison. The original SoTA models, such as DNABert and LOGO, use random span masking. We use the pretrained weights of the DNABert model, which was trained for 120K steps, and use it as a baseline model referred to as \textit{ORI 120K} model.

\noindent \textbf{Time-variant Masking Strategies.} We introduce one time-variant, curriculum masking-based baseline model called \textbf{CM-Step} where we start with the $100\%$ of random masking strategy (p\textsubscript{random}), with a uniform increase in the percentage of global PMI strategy (p\textsubscript{\globalgenemask} starting at $0\%$); p\textsubscript{random} + p\textsubscript{\globalgenemask} $= 100\%$. We first divide the maximum pretraining steps into blocks of 100 (as p\textsubscript{\globalgenemask} needs to increase from $0\%$ to $100\%$. We draw from a binomial distribution where the number of trials (n) is one and the probability of success (p) is p\textsubscript{\globalgenemask} for selecting the masking strategy.

\begin{table*}[!ht]
\centering
\setlength{\tabcolsep}{2pt}
\scalebox{0.96}{
\begin{tabular}{|p{0.95cm}p{1.5cm}p{1.2cm}|p{1.2cm}p{1.2cm}p{1.2cm}|p{1cm}p{1.5cm}p{1cm}p{1cm}|p{1cm}p{1.5cm}p{1cm}p{1cm}|} \hline
\textbf{Task} & \textbf{Dataset} & \multicolumn{12}{c|}{\textbf{Model}} \\
 &  & \multicolumn{4}{c}{\textbf{State-of-the-art models}} & \multicolumn{4}{c}{\textbf{DNABert}} & \multicolumn{4}{c|}{\textbf{LOGO}} \\ 
  &  & \textbf{DNABert-2 120K (BPE)$\bigstar$} & \textbf{DNABert 120K$\diamondsuit$ } & \textbf{NT-500M-Human$\diamondsuit$} & \textbf{DNABert-2 120K (k-mer)$\bigstar$} & \textbf{Global 10K} & \textbf{GeneMask 10K } & \textbf{CM-Step 10K} & \textbf{\dyngenemask{} 10K} & \textbf{Global 10K} & \textbf{GeneMask 10K} & \textbf{CM-Step 10K} & \textbf{\dyngenemask{} 10K} \\ \hline \hline
\multicolumn{14}{|c|}{\textbf{Species: Human}} \\ 
PD & all & 85.57 & 90.48 & 87.71 & 83.78 & 89.93 & 89.50 & 90.48 & 89.29 & 85.82 & 82.93 & 85.88 & 84.76 \\
(Human) & no tata & 92.55 & 93.05 & 90.75 & 92.65 & 91.09 & 91.73 & 91.83 & 92.41 & 89.52 & 88.54 & 89.79 & 88.62 \\
 & tata & 60.85 & 61.56 & 78.07 & 57.75 & 76.25 & 79.76 & 81.12 & 77.47 & 68.24 & 69.79 & 69.31 & 72.21 \\ 
CPD & all & 66.28 & 68.90 & 63.45 & 74.91 & 70.56 & 68.54 & 71.65 & 72.09 & 69.16 & 64.13 & 67.34 & 63.66 \\
(Human) & notata & 67.99 & 70.47 & 64.82 & 69.23 & 70.87 & 70.01 & 72.13 & 70.18 & 68.91 & 67.29 & 66.84 & 66.23 \\
 & tata & 72.73 & 76.06 & 71.34 & 74.91 & 76.96 & 74.95 & 75.51 & 83.50 & 70.85 & 53.65 & 55.55 & 61.71 \\ 
TFP & 0 & 66.99 & 66.84 & 61.59 & 67.99 & 65.89 & 67.40 & 65.44 & 66.07 & 64.21 & 67.09 & 63.94 & 65.34 \\
(Human) & 1 & 70.98 & 70.14 & 66.75 & 67.06 & 71.14 & 69.81 & 68.35 & 68.95 & 69.47 & 67.85 & 67.49 & 69.75 \\
 & 2 & 61.40 & 61.03 & 53.58 & 59.45 & 57.66 & 59.80 & 58.61 & 57.66 & 53.37 & 55.31 & 53.83 & 55.31 \\
 & 3 & 55.10 & 51.89 & 42.95 & 50.24 & 46.80 & 51.26 & 47.65 & 51.41 & 40.20 & 42.48 & 44.70 & 40.49 \\
 & 4 & 71.31 & 70.97 & 60.81 & 72.80 & 74.10 & 76.15 & 73.22 & 72.60 & 70.32 & 70.54 & 68.65 & 69.98 \\ 
Splice & Reconstruct & 79.62 & 84.07 & 79.71 & 77.90 & 83.02 & 84.84 & 84.74 & 84.12 & 77.92 & 74.01 & 80.25 & 75.20 \\ 
\multicolumn{2}{|c}{\textbf{Mean (Human)}} & 70.95 & \textbf{72.12} & 68.46 & \multicolumn{1}{l|}{70.72} & 72.86 & 73.65 & 73.39 & \textbf{73.81} & \textbf{69.00} & 66.97 & 67.80 & 67.77 \\ 
\hline \hline
\multicolumn{14}{|c|}{\textbf{Species: Non-human}} \\ 
EMP & H3 & 77.08 & 73.10 & 69.67 & 74.62 & 71.45 & 73.28 & 74.07 & 74.35 & 64.72 & 60.90 & 61.91 & 61.49 \\
(Yeast) & H3K14ac & 55.60 & 40.06 & 33.55 & 42.71 & 38.75 & 40.73 & 40.27 & 41.28 & 30.38 & 32.55 & 33.34 & 29.49 \\
 & H3K36me3 & 57.25 & 47.25 & 44.14 & 47.26 & 44.11 & 45.42 & 46.06 & 46.83 & 38.94 & 39.26 & 38.52 & 35.92 \\
 & H3K4me1 & 45.51 & 41.44 & 37.15 & 39.66 & 41.63 & 42.36 & 40.91 & 44.61 & 31.19 & 28.66 & 31.04 & 25.38 \\
 & H3K4me2 & 40.83 & 32.27 & 30.87 & 25.33 & 30.60 & 33.14 & 31.40 & 33.43 & 30.99 & 29.32 & 30.11 & 27.11 \\
 & H3K4me3 & 42.57 & 27.81 & 24.00 & 27.43 & 25.91 & 25.92 & 24.93 & 30.24 & 18.34 & 15.35 & 22.65 & 12.22 \\
 & H3K79me3 & 66.01 & 61.17 & 58.35 & 61.03 & 59.18 & 60.45 & 59.20 & 59.83 & 54.36 & 52.19 & 55.70 & 53.38 \\
 & H3K9ac & 56.79 & 51.22 & 45.81 & 49.35 & 49.24 & 52.22 & 51.77 & 52.77 & 43.54 & 42.16 & 45.62 & 40.04 \\
 & H4 & 80.07 & 79.26 & 76.17 & 78.61 & 76.38 & 76.04 & 75.83 & 76.66 & 72.81 & 66.51 & 71.51 & 68.85 \\
 & H4ac & 54.19 & 37.43 & 33.74 & 37.14 & 33.89 & 37.43 & 35.69 & 36.21 & 27.76 & 27.84 & 31.84 & 27.5 \\ 
TFP & 0 & 48.01 & 44.42 & 31.04 & 48.96 & 49.48 & 52.57 & 50.48 & 54.32 & 12.02 & 27.93 & 27.16 & 42.23 \\
(Mouse) & 1 & 81.86 & 78.94 & 75.04 & 81.69 & 79.70 & 79.05 & 79.90 & 80.51 & 71.30 & 69.73 & 70.93 & 69.50 \\
 & 2 & 82.98 & 71.44 & 61.67 & 81.71 & 75.50 & 78.08 & 74.40 & 80.70 & 52.50 & 59.80 & 55.57 & 78.19 \\
 & 3 & 73.22 & 44.89 & 29.17 & 63.17 & 51.00 & 60.27 & 52.51 & 61.01 & 34.87 & 48.96 & 34.38 & 61.53 \\
 & 4 & 46.15 & 42.48 & 29.27 & 42.83 & 41.04 & 42.60 & 42.68 & 44.94 & 21.40 & 28.78 & 25.89 & 26.60 \\ 
\multicolumn{2}{|c}{\textbf{Mean (Non-human)}} & 60.54 & 51.55 & 45.31 & \multicolumn{1}{l|}{\textbf{53.43}} & 51.19 & 53.30 & 52.01 & \textbf{54.51} & 40.34 & 42.00 & 42.41 & \textbf{43.96} \\
\hline
\end{tabular}}
\label{tab:gue-full-data-compare}
\caption{Performance comparison in terms of Matthews Correlation Coefficient (MCC) metric on the Genome Understanding Evaluation (GUE) benchmark. However, for the masking baselines, we pretrain for 10000 steps, following the evaluation setup prescribed by~\citet{royecai2023}. The task names are shortened due to space constraints - Promoter Detection (PD), Core Promoter Detection (CPD), Transcription Factor Prediction (TFP),  Splice Site Prediction (Splice) and Epigenetic Marks Prediction (EMP). $\diamondsuit$ and $\bigstar$ indicate that the values are directly taken from Tables 6 and 10 of~\citet{zhou2024dnabert}. The statistical significance test results using paired t-test are provided in Table~\ref{tab:gue-benchmark-stat}}
\end{table*}

\section{Experimental Results}
We show the performance comparison of our proposed masking algorithm, \dyngenemask{} in Table~\ref{tab:gue-full-data-compare} with the SoTA gene foundational models (DNABert-2, Nucleotide Transformer), the SoTA masking strategy for gene transformers (\genemask), and few baseline masking approaches. \dyngenemask{} outperforms \genemask{} in both the full dataset setting on the GUE benchmark (see RQ1) as well as the few-shot setup (see RQ3). In this paper, we explore the following research questions:

\noindent \textbf{RQ1: \dyngenemask{} evaluation on the full dataset with GUE benchmark.}
We observe that our proposed curriculum-trained model, \textit{DNABert \dyngenemask{} 10K}, achieves the best performance in both Human and Non-human tasks, outperforming both the k-mer-based SoTA models as well as the best masking strategy-based baseline model (GM 10K). \textit{LOGO \dyngenemask{} 10K}, a 15 million parameter model (compared to 117 million parameters of DNABert-2, achieves 93.97\% of the best SoTA model's performance (based on \textit{Matthews Correlation Coefficient} score), using just 10K steps instead of 120K. Table~\ref{tab:gue-benchmark-stat} shows the statistical significance test results where we observe that the performance improvement of \dyngenemask{} over DNABert 120K and DNABert-2 120K is statistically significant, unlike \genemask{} that fails to outperform DNABert 120K by a significant margin. Although \dyngenemask{} improves over \genemask{} by $1.20\%$, the improvement is not statistically significant.

\begin{table}[t]
    \centering
    \begin{tabular}{ccc}
    \hline
        \textbf{Models} & \textbf{GUE Dataset} & \textbf{p-value}\\ \hline 
        DNABert 120K &  $60.69 \pm 17.72$&0.021\\
        DNABert-2 120K (k-mer) &  $61.12 \pm 17.77$&0.043\\
        \genemask{} 10K&  $62.34 \pm 17.69$&0.112\\ 
        \dyngenemask{} 10K &  $63.09 \pm 17.49$&\\ \hline
    \end{tabular}
    \label{tab:gue-benchmark-stat}
        \caption{Performance comparison of \dyngenemask{} with state-of-the-art models on complete GUE benchmark. Statistical significance results using `paired t-test' are also provided. p-value $\leq 0.05$ indicates that the performance improvement of \dyngenemask{} is statistically significant}
\end{table}

\begin{figure}[b]
    \centering
    \includegraphics[width=0.4\textwidth]{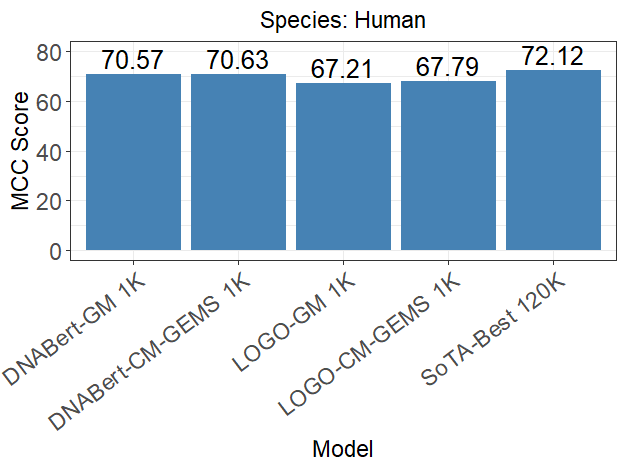} \hfill
    \includegraphics[width=0.4\textwidth]{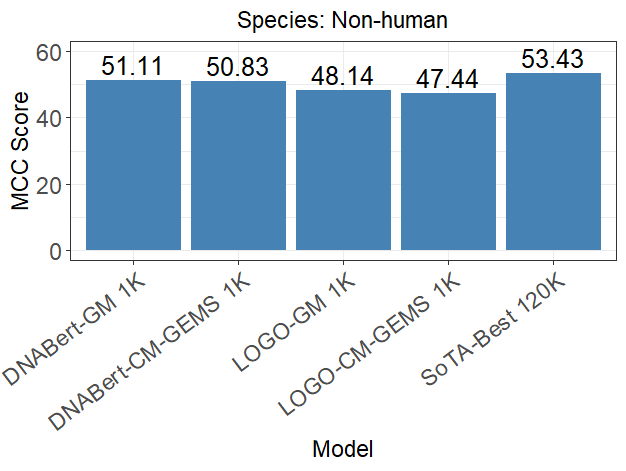}
    \label{fig:gue1K}
            \caption{Performance comparison on GUE benchmark at reduced compute 1K pretraining steps and reduced model size (LOGO). (left) Human species, (right) Non-human species. The SoTA models are pretrained for 120K steps on the Human Reference Genome}
\end{figure}

\noindent \textbf{RQ2: \dyngenemask{} evaluation at reduced compute and model size with GUE benchmark data.} We evaluate the effectiveness of our model trained with curriculum masking by only pretraining for $10\%$ steps (10\% of 10000 = 1000 steps), i.e., at just 1000 steps. Figure~\ref{fig:gue1K} shows the performance comparison of two such settings: (i) we keep the same architecture, which is DNABert, and (ii) we reduce the number of transformer blocks (12 to 2) by evaluating with the LOGO model. Table~\ref{tab:gue-1K-stats} of the Appendix provides individual task-wise performance on the GUE benchmark.
We observe that \dyngenemask{} 1K outperforms \genemask{} 1K models in the case of DNABert and LOGO for the `Human' species-related tasks of the GUE benchmark data. However, in the case of tasks related to the `Non-human' species, we observe a reverse trend.

We also explore the perspective of `intrinsic dimensionality' used in recent works~\citep{aghajanyan-etal-2021-intrinsic,zhang-etal-2023-fine} to better understand fine-tuning phenomena in our case. \textit{Intrinsic dimensionality} is defined as the minimum number of model parameters required to achieve $90\%$ of the original model performance. Based on Figure~\ref{fig:gue1K}, we observe that in the case of both Human and Non-human species, \dyngenemask{} achieves $90\%$ of the SoTA-best 120K model, except for LOGO 1K for Non-human species, i.e., it satisfies the criterion for intrinsic dimensionality. This brings up an interesting research direction for building smaller, energy-efficient models that achieve similar performance.

\begin{table*}[!ht]
    \centering
    \scalebox{0.79}{
    \addtolength{\tabcolsep}{-0.4em}
    \begin{tabular}{|p{2cm}ccccccccccp{0.8cm}|}
            \hline 
          \textbf{Model Type} &  \multicolumn{2}{c}{\textbf{Prom-core}}&\multicolumn{2}{c}{\textbf{Prom-300}} & \multicolumn{2}{c}{\textbf{Cohn-enh}}& \multicolumn{2}{c}{\textbf{Splice-40}}& \multicolumn{2}{c}{\textbf{Sil-300}} & \textbf{Mean}\\
            &  \textbf{10}&  \textbf{50}&  \textbf{10}&\textbf{50}& \textbf{10}& \textbf{50}& \textbf{10}& \textbf{50}& \textbf{10}& \textbf{50} & \\ \hline \hline
    DNABert 120K& $0.606 \pm 0.045$ & $0.687 \pm 0.024$ & $0.638 \pm 0.070$ & $0.808 \pm 0.019$& $0.582 \pm 0.030$ & $0.638 \pm 0.020$ & $0.404 \pm 0.019$& $0.472 \pm 0.048$ & $0.711 \pm 0.062$ & $0.849 \pm 0.020$ & 0.640\\ 
    DNABert 10K& $0.586 \pm 0.051$ & $0.653 \pm 0.058$ & $0.601 \pm 0.065$ & $0.789 \pm 0.059$& $0.579 \pm 0.047$& $0.634 \pm 0.031$& $0.409 \pm 0.017$ & $0.512 \pm 0.014$& $0.688 \pm 0.068$& $0.823 \pm 0.031$& 0.627\\
   GM 10K& $0.602 \pm 0.058$ & $0.678 \pm 0.026$ & $0.625 \pm 0.09$ & $0.781 \pm 0.097$ & $0.622 \pm 0.050$& $0.648 \pm 0.016$& $0.392 \pm 0.02$& $0.505 \pm 0.018$ & $0.715 \pm 0.061$& $0.818 \pm 0.029$& 0.639\\ \hline \hline
   Global 10K&  $0.605 \pm 0.05$&  $0.691 \pm 0.022$& $0.664 \pm 0.076$ &  $0.834 \pm 0.024$& $0.605 \pm 0.047$ & $0.627 \pm 0.04$ & $0.406 \pm 0.021$ & $0.514 \pm 0.014$& $0.768 \pm 0.062$& $0.849 \pm 0.011$& \textbf{0.656}\\ 
   CM-Step 10K& $0.579 \pm 0.049$ &  $0.675 \pm 0.022$&$0.649 \pm 0.091$&$0.799 \pm 0.034$&$0.608 \pm 0.043$&$0.649 \pm 0.017$&$0.402 \pm 0.02$&$0.557 \pm 0.022$&$0.706 \pm 0.044$&$0.835 \pm 0.023$& 0.646\\ 
   \dyngenemask{} 10K& $0.594 \pm 0.052$ &  $0.684 \pm 0.023$&$0.629 \pm 0.075$&$0.822 \pm 0.02$&$0.6 \pm 0.059$&$0.645 \pm 0.033$&$0.394 \pm 0.018$&$0.516 \pm 0.019$&$0.714 \pm 0.057$&$0.844 \pm 0.012$& 0.644\\
\hline
    \end{tabular}}
    \label{tab:fewshot-perf}
    \caption{Few-shot evaluation of \dyngenemask{} and baseline masking strategies in terms of accuracy across 10 and 50-shot settings with the base model being DNABert~\citep{DNABert}; the evaluation setup is taken from ~\citet{royecai2023}, with the addition of the silencer task. We observe that \dyngenemask{} 10K outperforms GM 10K and the DNABert 120K model; Global 10K achieves the best performance.}
\end{table*}
\vspace{3em}

\noindent \textbf{RQ3: \dyngenemask{} evaluation in few-shot setting.} 
We show the few-shot evaluation results in Table~\ref{tab:fewshot-perf}. Here, we observe that our proposed curriculum-trained model, \dyngenemask{} outperforms the current state-of-art masking strategy for gene transformers called \textsc{GeneMask} (GM 10K) in terms of mean accuracy over five datasets by a small margin of $0.78\%$ ($0.639$ versus $0.644$). We also observe that the Global 10K model achieved the best performance (mean accuracy of $0.656$) among the baseline models, highlighting that such an only-hard-masking strategy proves useful in few-shot settings. However, based on our observation of Global 10K on the full dataset setting on the GUE benchmark as evident in Table~\ref{tab:gue-full-data-compare}, Global 10K performs poorly, we thus conclude that the efficacy of such a masking scheme is limited to few-shot settings.

\begin{figure}[b]
    \centering
\includegraphics[width=\columnwidth]{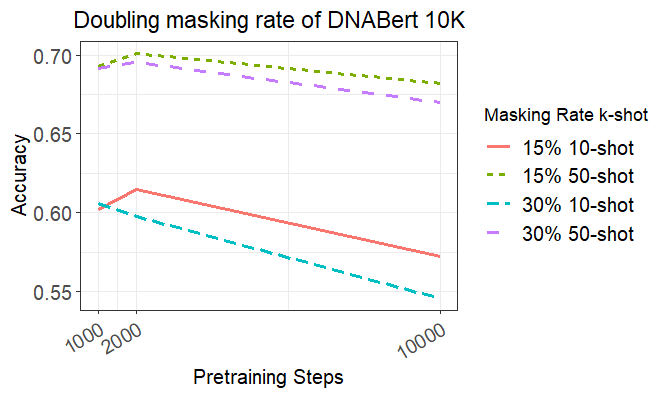}
    \label{fig:masking-rate}
            \caption{Impact of doubling the masking rate of DNABert 10K model. The performance stays almost the same at 1K steps, but then it slowly decreases at 2K and 10K steps} 
\end{figure}

\noindent \textbf{RQ4: Impact of doubling the masking rate of DNABert model in a few-shot setting.}~\citet{wettig-etal-2023-mask} showed in the NLP domain that both Random Span masking and PMI masking achieve the best downstream task performance (in both GLUE and SQuAD benchmark datasets) when the masking rate is kept at $30\%$ (please see Figure 5 of~\citet{wettig-etal-2023-mask}). Therefore, we test the hypothesis of whether trivially doubling the masking rate leads to a performance improvement of the same order as that achieved by PMI-based masking strategies. We chose a few-shot setting (10 and 50-shot) for this purpose to notice the impact of this change in pretraining configuration by limiting the effect of huge amounts of fine-tuning data. Figure~\ref{fig:masking-rate} shows that at the early stages of pretraining (i.e., at 1000 steps), there is a marginal difference, but as we progress further (at 2000 and 10000 steps), the performance (mean accuracy over five datasets) of DNABert with double the masking rate goes on decreasing. The performance drop at 10K steps over the standard masking rate of $15\%$ is $4.71\%$ and $1.79\%$ respectively. The reason may be that with a higher masking rate, the model sees a much-reduced amount of unmasked or actual tokens that hinder the learning process. 

\vspace{3em}

%% file: conclusion.tex
\section{Conclusion}
We develop a novel (time-variant, easy-to-hard) curriculum masking algorithm,~\dyngenemask{} for optimal masked language modeling training of gene transformer models.  We also develop a time-invariant masking strategy called \globalgenemask{}, where we specifically prioritize the top-ranked PMI tokens, which we refer to as the `hard' strategy of our proposed curriculum masking strategy. Our study highlights the limitations of conventional tokenization methods in gene transformer models and proposes a novel curriculum masking approach to address these shortcomings. By systematically increasing the difficulty of the masked token prediction task based on Pointwise Mutual Information, we enhance the representation learning capabilities of Gene Transformer models such as DNABert and LOGO. \dyngenemask{} outperforms state-of-the-art models on average, in both few-shot gene sequence classification tasks (5 datasets) as well as on the Genomic Understanding Evaluation benchmark (27 datasets) trained on the full dataset. We also show results on the full dataset (GUE benchmark) and observe that \dyngenemask{} at 1K steps and \globalgenemask{} at 10K can achieve performance within a $90\%$ margin than state-of-the-art models such as DNABert-2 and Nucleotide transformers trained for 120K steps. This shows the potential of adaptive masking strategies for fast pretraining and effectively reducing the compute requirement for training gene foundational models, a major limitation of current models as pointed out by~\citet{consens2023transformers}.

\noindent \textbf{Future Work.} Given the significant performance gain observed for DNA foundational models due to intelligent (PMI-based) masking strategies, we will extend our work to other foundational models based on genomics data, specifically RNA-based models such as CodonBERT~\citep{codonbert}, RNABERT~\citep{rnabert2022} and single-cell RNA-based models such as GeneFormer~\citep{geneformer2023}. Although PMI indirectly captures DNA sequence motifs~\citep{royecai2023}, a much-needed inter-disciplinary research direction is to involve more biologically grounded pretraining or fine-tuning objectives instead of MLM~\citep{consens2023transformers}.  We will perform a more systematic exploration of the `intrinsic dimensionality' to better understand the fine-tuning dynamics of gene transformer models; to the best of our knowledge, this is a vastly unexplored domain with very limited research. We make the codebase available at \url{https://github.com/roysoumya/curriculum-GeneMask}. 

%% file: appendix.tex
\appendix

\section{Research Background}
The formal definition of Point-wise Mutual Information (PMI\textsubscript{k}), where k $>$ 2, as proposed by~\citet{pmi-masking}. 

\begin{equation}\label{eq:pmi}
	\textrm{PMI}_k(w_1\ldots w_k)=\min_{\sigma\in\textrm{seg}(w_1\ldots w_k)}\log\frac{p(w_1\ldots w_k)}{\prod_{s\in\sigma}p(s)}
	\end{equation}
\raggedbottom
Here, $\textrm{seg}(w_1\ldots w_k)$ represents the set of all continuous segmentations of the $k$-mer ``$w_1\ldots w_k$" (identity segmentation is not included). 

\section{Experimental Setup}\label{sec:expt-setup-expt-setup}
\subsection{Dataset Construction}
The silencer dataset is obtained from the `Home Sapiens' annotation file from the \textit{SilencerDB} database, available at \url{http://health.tsinghua.edu.cn/SilencerDB/download/Species/Homo_sapiens.fa}.

\subsection{Training Details}\label{sec:training-details-app}
\noindent \textbf{Finetuning parameter configuration.} The models are finetuned for $20$ epochs at a learning rate of $4e^{-4}$, warmup steps percentage of $10\%$, hidden dropout probability of $0.1$, weight decay as $0.01$, per GPU train batch size as $5$ and use the \textit{AdamW} optimizer for the 10, 50 and 100-shot setting. However, the performance drops in 500 and 1000-shot settings due to overfitting. Therefore, for the 500-shot settings and above, we use the same hyperparameters as the original DNABert paper --- a lower learning rate from $4e^{-4}$ to $5e^{-5}$ and a lower number of epochs from 20 to 5. Mosbach et al.~\cite{mosbach2021} observe that such a high number of fine-tuning epochs helps address random initialization issues in the low-resource settings (10, 50-shot). 

\begin{table}[!ht]
    \centering
    \begin{tabular}{ccc}
    \hline
        \textbf{Parameter} & \textbf{DNABert} & \textbf{LOGO} \\
        \hline
        Hidden Size & 768 & 256 \\
        Hidden Layers & 12 & 2 \\
        Attention Heads & 12 & 8 \\
       Per GPU train batch size & 10 & 5 \\
        Hidden Dropout Probability & 0.1 & 0 \\
        Attention Dropout Probability & 0.1 & 0 \\
        Intermediate Size & 3072 &  3072 \\
        Embedding Size & 512 & 512 \\ \hline
    \end{tabular}
    \label{tab:dna-logo-config}
            \caption{Difference between parameters of DNABert and LOGO}
\end{table}


\begin{table*}[t]
\centering
\label{tab:gue-1K-stats}
\setlength{\tabcolsep}{2pt}
\scalebox{1.1}{
\begin{tabular}{|p{0.95cm}p{1.5cm}p{1.2cm}|p{1.2cm}p{1.2cm}p{1.2cm}|p{1.5cm}p{1cm}p{1cm}|p{1.5cm}p{1cm}p{1cm}|} 
\hline
\textbf{Task} & \textbf{Dataset} & \multicolumn{10}{c|}{\textbf{Model}} \\
 &  & \multicolumn{4}{c}{\textbf{State-of-art-models}} & \multicolumn{3}{c}{\textbf{DNABert}} & \multicolumn{3}{c|}{\textbf{LOGO}} \\ 
 &  & \textbf{DNABert-2 120K (BPE)$\bigstar$} & \textbf{DNABert 120K$\diamondsuit$ } & \textbf{NT-500M-Human$\diamondsuit$} & \textbf{DNABert-2 120K (kmer)$\bigstar$} & \textbf{GeneMask 1K} & \textbf{CM-Step 1K} & \textbf{CM-GEMS 1K} & \textbf{GeneMask 1K} & \textbf{CM-Step 1K} & \textbf{CM-GEMS 1K} \\ \hline \hline
\multicolumn{12}{|c|}{\textbf{Species: Human}} \\ 
PD & all & 85.57 & 90.48 & 87.71 & 83.78 & 85.70 & 86.66 & 87.05 & 84.76 & 84.44 & 84.70 \\
(Human) & no tata & 92.55 & 93.05 & 90.75 & 92.65 & 89.59 & 89.50 & 89.14 & 89.32 & 89.23 & 89.30 \\
 & tata & 60.85 & 61.56 & 78.07 & 57.75 & 64.81 & 69.06 & 68.46 & 64.43 & 70.62 & 64.41 \\ 
CPD & all & 66.28 & 68.90 & 63.45 & 74.91 & 65.93 & 65.98 & 66.02 & 62.77 & 63.72 & 63.43 \\
(Human) & notata & 67.99 & 70.47 & 64.82 & 69.23 & 67.00 & 67.03 & 66.91 & 66.31 & 67.52 & 66.80 \\
 & tata & 72.73 & 76.06 & 71.34 & 74.91 & 69.74 & 64.35 & 74.10 & 66.08 & 61.53 & 70.74 \\ 
TFP & 0 & 66.99 & 66.84 & 61.59 & 67.99 & 66.33 & 65.34 & 63.03 & 63.56 & 61.52 & 62.97 \\
(Human) & 1 & 70.98 & 70.14 & 66.75 & 67.06 & 70.11 & 69.79 & 71.04 & 69.89 & 69.39 & 69.11 \\
 & 2 & 61.40 & 61.03 & 53.58 & 59.45 & 63.35 & 61.68 & 61.97 & 53.27 & 53.24 & 55.88 \\
 & 3 & 55.10 & 51.89 & 42.95 & 50.24 & 48.80 & 47.42 & 46.04 & 40.66 & 41.09 & 38.57 \\
 & 4 & 71.31 & 70.97 & 60.81 & 72.80 & 75.63 & 73.03 & 75.32 & 70.29 & 72.68 & 72.80 \\ 
Splice & Reconstruct & 79.62 & 84.07 & 79.71 & 77.90 & 78.79 & 75.99 & 78.52 & 75.12 & 75.55 & 74.71 \\ 
\multicolumn{2}{|c}{\textbf{Mean (Human)}} & 70.95 & \textbf{72.12} & 68.46 & \multicolumn{1}{l|}{70.72} & 70.57 & 69.65 & \multicolumn{1}{l|}{\textbf{70.63}} & 67.21 & 67.54 & \textbf{67.79} \\ 
\hline \hline
\multicolumn{12}{|c|}{\textbf{Species: non-Human}} \\ 
EMP & H3 & 77.08 & 73.10 & 69.67 & 74.62 & 69.25 & 69.66 & 67.51 & 66.34 & 67.76 & 66.26 \\
(Yeast) & H3K14ac & 55.60 & 40.06 & 33.55 & 42.71 & 39.66 & 38.70 & 38.35 & 39.08 & 34.95 & 35.75 \\
 & H3K36me3 & 57.25 & 47.25 & 44.14 & 47.26 & 40.87 & 43.01 & 40.31 & 41.73 & 40.39 & 40.15 \\
 & H3K4me1 & 45.51 & 41.44 & 37.15 & 39.66 & 33.13 & 35.44 & 32.12 & 31.38 & 30.69 & 30.97 \\
 & H3K4me2 & 40.83 & 32.27 & 30.87 & 25.33 & 29.81 & 30.03 & 31.82 & 30.10 & 29.19 & 30.35 \\
 & H3K4me3 & 42.57 & 27.81 & 24.00 & 27.43 & 27.35 & 26.45 & 24.05 & 23.48 & 21.62 & 23.02 \\
 & H3K79me3 & 66.01 & 61.17 & 58.35 & 61.03 & 59.98 & 59.54 & 61.05 & 58.45 & 56.53 & 58.00 \\
 & H3K9ac & 56.79 & 51.22 & 45.81 & 49.35 & 47.13 & 46.97 & 45.12 & 46.20 & 44.39 & 45.40 \\
 & H4 & 80.07 & 79.26 & 76.17 & 78.61 & 74.05 & 74.84 & 76.68 & 72.09 & 72.99 & 71.61 \\
 & H4ac & 54.19 & 37.43 & 33.74 & 37.14 & 35.94 & 36.65 & 35.09 & 32.52 & 33.47 & 31.80 \\ 
TFP & 0 & 48.01 & 44.42 & 31.04 & 48.96 & 52.35 & 53.33 & 53.60 & 46.23 & 43.79 & 39.30 \\
(Mouse) & 1 & 81.86 & 78.94 & 75.04 & 81.69 & 75.91 & 76.32 & 75.42 & 70.14 & 71.16 & 70.44 \\
 & 2 & 82.98 & 71.44 & 61.67 & 81.71 & 72.63 & 75.07 & 76.83 & 72.14 & 73.22 & 74.75 \\
 & 3 & 73.22 & 44.89 & 29.17 & 63.17 & 65.80 & 65.29 & 64.12 & 62.36 & 58.19 & 61.71 \\
 & 4 & 46.15 & 42.48 & 29.27 & 42.83 & 42.76 & 39.89 & 40.45 & 29.79 & 35.02 & 32.03 \\ 
\multicolumn{2}{|c}{\textbf{Mean (non-Human)}} & 60.54 & 51.55 & 45.31 & \multicolumn{1}{l|}{\textbf{53.43}} & 51.11 & \textbf{51.41} & \multicolumn{1}{l|}{50.83} & \textbf{48.14} & 47.56 & 47.44 \\
 \hline
\end{tabular}}
\caption{Performance comparison in terms of Matthews Correlation Coefficient (MCC) metric on the GUE benchmark datasets that target the `Human' species. For the masking baselines, we pretrain for 1000 steps. For a fair comparison, we select variants of SoTA models that are trained only on the Human Reference Genome dataset and use the k-mer tokenization scheme (k=6); for DNABert-2, we use the results presented in Table 10 of Appendix of ~\citet{zhou2024dnabert}. The task names are shortened due to space constraints - Promoter Detection (PD), Core Promoter Detection (CPD), Transcription Factor Prediction (TFP), and Splice Site Prediction (Splice). $\diamondsuit$ and $\bigstar$ indicate that the values are directly taken from Tables 6 and 10 of~\citet{zhou2024dnabert}.}
\end{table*}

%% file: m2909.bbl
\begin{thebibliography}{36}
\providecommand{\natexlab}[1]{#1}
\providecommand{\url}[1]{\texttt{#1}}
\expandafter\ifx\csname urlstyle\endcsname\relax
  \providecommand{\doi}[1]{doi: #1}\else
  \providecommand{\doi}{doi: \begingroup \urlstyle{rm}\Url}\fi

\bibitem[Aghajanyan et~al.(2021)Aghajanyan, Gupta, and Zettlemoyer]{aghajanyan-etal-2021-intrinsic}
A.~Aghajanyan, S.~Gupta, and L.~Zettlemoyer.
\newblock Intrinsic dimensionality explains the effectiveness of language model fine-tuning.
\newblock In \emph{Proceedings of the 59th Annual Meeting of the Association for Computational Linguistics and the 11th International Joint Conference on Natural Language Processing (Volume 1: Long Papers)}, pages 7319--7328, Online, Aug. 2021.
\newblock \doi{10.18653/v1/2021.acl-long.568}.

\bibitem[Akiyama and Sakakibara(2022)]{rnabert2022}
M.~Akiyama and Y.~Sakakibara.
\newblock {Informative RNA base embedding for RNA structural alignment and clustering by deep representation learning}.
\newblock \emph{NAR Genomics and Bioinformatics}, 4\penalty0 (1):\penalty0 lqac012, 02 2022.
\newblock ISSN 2631-9268.
\newblock \doi{10.1093/nargab/lqac012}.

\bibitem[Anagnostidis et~al.(2023)Anagnostidis, Bachmann, and Hofmann]{anagnostidis2023navigating}
S.~Anagnostidis, G.~Bachmann, and T.~Hofmann.
\newblock Navigating scaling laws: Accelerating vision transformer's training via adaptive strategies.
\newblock \emph{arXiv preprint arXiv:2311.03233}, 2023.

\bibitem[Avsec et~al.(2021)Avsec, Agarwal, et~al.]{enformer}
{\v{Z}}.~Avsec, V.~Agarwal, et~al.
\newblock Effective gene expression prediction from sequence by integrating long-range interactions.
\newblock \emph{Nature Methods}, 18\penalty0 (10):\penalty0 1196--1203, Oct 2021.
\newblock ISSN 1548-7105.
\newblock \doi{10.1038/s41592-021-01252-x}.

\bibitem[Babjac et~al.(2023)Babjac, Lu, and Emrich]{codonbert}
A.~N. Babjac, Z.~Lu, and S.~J. Emrich.
\newblock Codonbert: Using bert for sentiment analysis to better predict genes with low expression.
\newblock In \emph{Proceedings of the 14th ACM International Conference on Bioinformatics, Computational Biology, and Health Informatics}, BCB '23, 2023.
\newblock ISBN 9798400701269.
\newblock \doi{10.1145/3584371.3613013}.

\bibitem[Badirli et~al.(2021)Badirli, Akata, et~al.]{Badirli2021}
S.~Badirli, Z.~Akata, et~al.
\newblock Fine-grained zero-shot learning with dna as side information.
\newblock In \emph{Advances in Neural Information Processing Systems}, volume~34, pages 19352--19362, 2021.

\bibitem[Consens et~al.(2023)Consens, Dufault, Wainberg, et~al.]{consens2023transformers}
M.~E. Consens, C.~Dufault, M.~Wainberg, et~al.
\newblock To transformers and beyond: Large language models for the genome.
\newblock \emph{arXiv preprint arXiv:2311.07621}, 2023.

\bibitem[Consortium(2019)]{human-reference}
G.~R. Consortium.
\newblock Genome reference consortium human build 38 patch release 13 (grch38.p13), 2019.
\newblock URL \url{https://www.ncbi.nlm.nih.gov/assembly/GCF_000001405.39/}.

\bibitem[Dalla-Torre et~al.(2023)Dalla-Torre, Gonzalez, et~al.]{dalla2023nucleotide}
H.~Dalla-Torre, L.~Gonzalez, et~al.
\newblock The nucleotide transformer: Building and evaluating robust foundation models for human genomics.
\newblock \emph{bioRxiv}, pages 2023--01, 2023.

\bibitem[Devlin et~al.(2019)Devlin, Chang, et~al.]{bert}
J.~Devlin, M.-W. Chang, et~al.
\newblock {BERT}: Pre-training of deep bidirectional transformers for language understanding.
\newblock In \emph{Proceedings of the 2019 Conference of the North {A}merican Chapter of the Association for Computational Linguistics: Human Language Technologies, Volume 1 (Long and Short Papers)}, pages 4171--4186, Minneapolis, Minnesota, June 2019.
\newblock \doi{10.18653/v1/N19-1423}.

\bibitem[Ji et~al.(2021)Ji, Zhou, et~al.]{DNABert}
Y.~Ji, Z.~Zhou, et~al.
\newblock {DNABERT: pre-trained Bidirectional Encoder Representations from Transformers model for DNA-language in genome}.
\newblock \emph{Bioinformatics}, 37\penalty0 (15):\penalty0 2112--2120, 02 2021.
\newblock ISSN 1367-4803.
\newblock \doi{10.1093/bioinformatics/btab083}.

\bibitem[Joshi et~al.(2020)Joshi, Chen, et~al.]{spanbert2020}
M.~Joshi, D.~Chen, et~al.
\newblock {SpanBERT: Improving Pre-training by Representing and Predicting Spans}.
\newblock \emph{Transactions of the Association for Computational Linguistics}, 8:\penalty0 64--77, 01 2020.
\newblock ISSN 2307-387X.
\newblock \doi{10.1162/tacl_a_00300}.

\bibitem[Lee et~al.(2022)Lee, Park, et~al.]{lee-etal-2022-efficient-pre}
M.~Lee, J.-H. Park, et~al.
\newblock Efficient pre-training of masked language model via concept-based curriculum masking.
\newblock In \emph{Proceedings of the 2022 Conference on Empirical Methods in Natural Language Processing}, pages 7417--7427, Abu Dhabi, United Arab Emirates, Dec. 2022.
\newblock \doi{10.18653/v1/2022.emnlp-main.502}.

\bibitem[Levine et~al.(2021)Levine, Lenz, et~al.]{pmi-masking}
Y.~Levine, B.~Lenz, et~al.
\newblock {\{}PMI{\}}-masking: Principled masking of correlated spans.
\newblock In \emph{International Conference on Learning Representations}, 2021.
\newblock URL \url{https://openreview.net/forum?id=3Aoft6NWFej}.

\bibitem[Madan et~al.(2024)Madan, Ristea, Nasrollahi, et~al.]{Madan2024WACV}
N.~Madan, N.-C. Ristea, K.~Nasrollahi, et~al.
\newblock Cl-mae: Curriculum-learned masked autoencoders.
\newblock In \emph{Proceedings of the IEEE/CVF Winter Conference on Applications of Computer Vision (WACV)}, pages 2492--2502, January 2024.

\bibitem[Mo et~al.(2021)Mo, Fu, et~al.]{genebert2021}
S.~Mo, X.~Fu, et~al.
\newblock Multi-modal self-supervised pre-training for regulatory genome across cell types.
\newblock \emph{CoRR}, abs/2110.05231, 2021.
\newblock URL \url{https://arxiv.org/abs/2110.05231}.

\bibitem[Mosbach et~al.(2021)Mosbach, Andriushchenko, et~al.]{mosbach2021}
M.~Mosbach, M.~Andriushchenko, et~al.
\newblock On the stability of fine-tuning {BERT:} misconceptions, explanations, and strong baselines.
\newblock In \emph{9th International Conference on Learning Representations, {ICLR} 2021}, 2021.
\newblock URL \url{https://openreview.net/forum?id=nzpLWnVAyah}.

\bibitem[Ng(2017)]{dnavec}
P.~Ng.
\newblock dna2vec: Consistent vector representations of variable-length k-mers.
\newblock \emph{CoRR}, abs/1701.06279, 2017.
\newblock URL \url{http://arxiv.org/abs/1701.06279}.

\bibitem[Nguyen et~al.(2023)Nguyen, Poli, Faizi, et~al.]{hyenadna}
E.~Nguyen, M.~Poli, M.~Faizi, et~al.
\newblock Hyenadna: Long-range genomic sequence modeling at single nucleotide resolution.
\newblock In \emph{Advances in Neural Information Processing Systems}, volume~36, pages 43177--43201, 2023.
\newblock URL \url{https://proceedings.neurips.cc/paper_files/paper/2023/file/86ab6927ee4ae9bde4247793c46797c7-Paper-Conference.pdf}.

\bibitem[Oubounyt et~al.(2019)Oubounyt, Louadi, Tayara, and Chong]{deepromoter2019}
M.~Oubounyt, Z.~Louadi, H.~Tayara, and K.~T. Chong.
\newblock Deepromoter: Robust promoter predictor using deep learning.
\newblock \emph{Frontiers in Genetics}, 10, 2019.
\newblock ISSN 1664-8021.
\newblock \doi{10.3389/fgene.2019.00286}.

\bibitem[Pavlova and Makhlouf(2023)]{pavlova-makhlouf-2023-bioptimus}
V.~Pavlova and M.~Makhlouf.
\newblock {BIO}ptimus: Pre-training an optimal biomedical language model with curriculum learning for named entity recognition.
\newblock In \emph{The 22nd Workshop on Biomedical Natural Language Processing and BioNLP Shared Tasks}, pages 337--349, Toronto, Canada, July 2023.
\newblock \doi{10.18653/v1/2023.bionlp-1.31}.

\bibitem[Roy et~al.(2023)Roy, Wallat, Sundaram, et~al.]{royecai2023}
S.~Roy, J.~Wallat, S.~S. Sundaram, et~al.
\newblock {GENEMASK:} fast pretraining of gene sequences to enable few-shot learning.
\newblock In \emph{{ECAI} 2023 - 26th European Conference on Artificial Intelligence, September 30 - October 4, 2023, Krak{\'{o}}w, Poland}, volume 372 of \emph{Frontiers in Artificial Intelligence and Applications}, pages 2002--2009, 2023.
\newblock \doi{10.3233/FAIA230492}.

\bibitem[Sadeq et~al.(2022)Sadeq, Xu, and McAuley]{sadeq-etal-2022-informask}
N.~Sadeq, C.~Xu, and J.~McAuley.
\newblock {I}nfor{M}ask: Unsupervised informative masking for language model pretraining.
\newblock In \emph{Proceedings of the 2022 Conference on Empirical Methods in Natural Language Processing}, pages 5866--5878, Abu Dhabi, United Arab Emirates, Dec. 2022.
\newblock URL \url{https://aclanthology.org/2022.emnlp-main.395}.

\bibitem[Schick and Sch{\"u}tze(2021{\natexlab{a}})]{schick-schutze-2021-exploiting}
T.~Schick and H.~Sch{\"u}tze.
\newblock Exploiting cloze-questions for few-shot text classification and natural language inference.
\newblock In \emph{Proceedings of the 16th Conference of the European Chapter of the Association for Computational Linguistics: Main Volume}, pages 255--269, Online, Apr. 2021{\natexlab{a}}.
\newblock \doi{10.18653/v1/2021.eacl-main.20}.

\bibitem[Schick and Sch{\"u}tze(2021{\natexlab{b}})]{schick-schutze-2021-just}
T.~Schick and H.~Sch{\"u}tze.
\newblock It{'}s not just size that matters: Small language models are also few-shot learners.
\newblock In \emph{Proceedings of the 2021 Conference of the North American Chapter of the Association for Computational Linguistics: Human Language Technologies}, pages 2339--2352, Online, June 2021{\natexlab{b}}.
\newblock \doi{10.18653/v1/2021.naacl-main.185}.

\bibitem[Sennrich et~al.(2016)Sennrich, Haddow, et~al.]{senrich:2016:acl:wholewordmasking}
R.~Sennrich, B.~Haddow, et~al.
\newblock Neural machine translation of rare words with subword units.
\newblock In \emph{Proceedings of the 54th Annual Meeting of the Association for Computational Linguistics, {ACL} 2016, Volume 1: Long Papers}, 2016.
\newblock \doi{10.18653/v1/p16-1162}.

\bibitem[Sun et~al.(2019)Sun, Wang, et~al.]{sun2019ernie}
Y.~Sun, S.~Wang, et~al.
\newblock Ernie: Enhanced representation through knowledge integration.
\newblock \emph{arXiv preprint arXiv:1904.09223}, 2019.

\bibitem[Theodoris et~al.(2023)Theodoris, Xiao, et~al.]{geneformer2023}
C.~V. Theodoris, L.~Xiao, et~al.
\newblock Transfer learning enables predictions in network biology.
\newblock \emph{Nature}, 618\penalty0 (7965):\penalty0 616--624, Jun 2023.
\newblock ISSN 1476-4687.
\newblock \doi{10.1038/s41586-023-06139-9}.

\bibitem[Wettig et~al.(2023)Wettig, Gao, Zhong, and Chen]{wettig-etal-2023-mask}
A.~Wettig, T.~Gao, Z.~Zhong, and D.~Chen.
\newblock Should you mask 15{\%} in masked language modeling?
\newblock In \emph{Proceedings of the 17th Conference of the European Chapter of the Association for Computational Linguistics}, pages 2985--3000, Dubrovnik, Croatia, May 2023.
\newblock \doi{10.18653/v1/2023.eacl-main.217}.

\bibitem[Yang et~al.(2023)Yang, Zhang, and Zhao]{yang-etal-2023-learning}
D.~Yang, Z.~Zhang, and H.~Zhao.
\newblock Learning better masking for better language model pre-training.
\newblock In \emph{Proceedings of the 61st Annual Meeting of the Association for Computational Linguistics (Volume 1: Long Papers)}, pages 7255--7267, Toronto, Canada, July 2023.
\newblock \doi{10.18653/v1/2023.acl-long.400}.

\bibitem[Yang et~al.(2022)Yang, Huang, et~al.]{yang:2021:bioarxiv:logo}
M.~Yang, L.~Huang, et~al.
\newblock {Integrating convolution and self-attention improves language model of human genome for interpreting non-coding regions at base-resolution}.
\newblock \emph{Nucleic Acids Research}, 50\penalty0 (14):\penalty0 e81--e81, 05 2022.
\newblock ISSN 0305-1048.
\newblock \doi{10.1093/nar/gkac326}.

\bibitem[Yasutomi and Ogata(2023)]{yasutomi2023}
A.~Y. Yasutomi and T.~Ogata.
\newblock Automatic action space curriculum learning with dynamic per-step masking.
\newblock In \emph{2023 IEEE 19th International Conference on Automation Science and Engineering (CASE)}, pages 1--7, 2023.
\newblock \doi{10.1109/CASE56687.2023.10260397}.

\bibitem[Zaheer et~al.(2020)Zaheer, Guruganesh, et~al.]{bigbird2020}
M.~Zaheer, G.~Guruganesh, et~al.
\newblock Big bird: Transformers for longer sequences.
\newblock In \emph{Advances in Neural Information Processing Systems}, volume~33, pages 17283--17297, 2020.

\bibitem[Zeng et~al.(2020)Zeng, Chen, Cui, Chen, Gao, and Jiang]{silencer2020}
W.~Zeng, S.~Chen, X.~Cui, X.~Chen, Z.~Gao, and R.~Jiang.
\newblock {SilencerDB: a comprehensive database of silencers}.
\newblock \emph{Nucleic Acids Research}, 49\penalty0 (D1):\penalty0 D221--D228, 10 2020.
\newblock ISSN 0305-1048.
\newblock \doi{10.1093/nar/gkaa839}.

\bibitem[Zhang et~al.(2023)Zhang, Liu, and Shao]{zhang-etal-2023-fine}
Z.~Zhang, B.~Liu, and J.~Shao.
\newblock Fine-tuning happens in tiny subspaces: Exploring intrinsic task-specific subspaces of pre-trained language models.
\newblock In \emph{Proceedings of the 61st Annual Meeting of the Association for Computational Linguistics (Volume 1: Long Papers)}, pages 1701--1713, Toronto, Canada, July 2023.
\newblock \doi{10.18653/v1/2023.acl-long.95}.

\bibitem[Zhou et~al.(2024)Zhou, Ji, Li, Dutta, Davuluri, and Liu]{zhou2024dnabert}
Z.~Zhou, Y.~Ji, W.~Li, P.~Dutta, R.~V. Davuluri, and H.~Liu.
\newblock {DNABERT}-2: Efficient foundation model and benchmark for multi-species genomes.
\newblock In \emph{The Twelfth International Conference on Learning Representations}, 2024.
\newblock URL \url{https://openreview.net/forum?id=oMLQB4EZE1}.

\end{thebibliography}
